\documentclass[a4paper, 10pt, conference]{ieeeconf}      
\usepackage{FG2020}
\usepackage{amssymb}
\FGfinalcopy 

\IEEEoverridecommandlockouts                              
\usepackage{bm}
\usepackage{graphicx}
\usepackage{amsmath}
\usepackage{dblfloatfix}
\usepackage{float}
\usepackage{multirow}
\usepackage{subfig}
\usepackage{array}

\title{\LARGE \bf
ReenactNet: Real-time Full Head Reenactment
}

\author{\parbox{16cm}{\centering
    {\large Mohammad Rami Koujan$^{*1,3}$, Michail Christos Doukas$^{*2,3}$, Anastasios Roussos$^{4,3}$, Stefanos Zafeiriou$^{2,3}$}\\
    {\normalsize
    $^{1}$College of Engineering, Mathematics and Physical Sciences, University of Exeter, UK\\  $^{2}$Department of Computing, Imperial College London, UK\\$^{4}$Institute of Computer Science (ICS), Foundation for Research and Technology - Hellas (FORTH), Greece\\ $^{3}$FaceSoft.io, London, UK}}
\thanks{$^{*}$ Equal contribution}
}

\begin{document}

\ifFGfinal
\thispagestyle{empty}
\pagestyle{empty}
\else
\author{Anonymous FG2020 submission\\ Paper ID \FGPaperID \\}
\pagestyle{plain}
\fi
\maketitle

\begin{abstract}
video-to-video synthesis is a challenging problem aiming at learning a translation function between a sequence of semantic maps and a photo-realistic video depicting the characteristics of a driving video. We propose a head-to-head system of our own implementation capable of fully transferring the human head 3D pose, facial expressions and eye gaze from a source to a target actor, while preserving the identity of the target actor.  Our system produces high-fidelity, temporally-smooth and photo-realistic synthetic videos faithfully transferring the human time-varying head attributes from the source to the target actor. Our proposed implementation: 1) works in real time ($\sim 20$ fps), 2) runs on a commodity laptop with a webcam as the only input, 3) is interactive, allowing the participant to drive a target person, e.g. a celebrity, politician, etc, instantly by varying their expressions, head pose, and eye gaze, and visualising the synthesised video concurrently.



\end{abstract}
\section{INTRODUCTION}
While facial reenactment aims at only transferring the expression between source and target actors, full-head reenactment transfers all the head time-varying features (3D pose, facial expressions, mouth movements, eyes gaze, etc.). This constitutes a challenging task with many applications in video editing, movie dubbing, telepresence and virtual reality.
The majority of facial reenactment methods transfer the expressions of the source actor by modifying the deformations within solely the inner facial region of the target actor, without altering the head movements of the target video. Thus, even in cases where this expression transfer is performed well, the overall reenactment result might seem uncanny and non-plausible, as the head motion of the target may not match with the transferred expressions. 
Our proposed approach overcomes all the aforementioned limitations. We fully transfer the pose, facial expression, eye gaze movement from a source to a target video, while preserving the identity of the target and maintaining a consistent motion of the upper body part. Given that people easily detect mistakes in the appearance of a human face (uncanny valley effect), we give specific attention in the details of the mouth and teeth. Our system generates photo-realistic and temporally consistent videos of faces. The proposed \textbf{head2head} pipeline consists of two successive stages: 1) 3D facial reconstruction and tracking, followed by 2) learning-based video rendering, using a neural network. Please check our recently accepted paper in FG2020 for more details \cite{mrkoujan}.\\
1- \textbf{Facial Reconstruction and Tracking}. The aim of this stage is to estimate and track a set of facial parameters in the input video representing: 1) identity, as well as the per-frame 2) facial expressions, 3) eye gaze movements, and 4) 3D pose. We harness the power of 3DMMs for 3D reconstructing and tracking the faces  and encode the estimated parameters in a constructive representation aiding the conditioning of our learning-based video renderer. More specifically, during training we extract the facial parameters from the target video and render the per-frame reconstructed 3D face with a fixed color for each rendered vertex. We call this representation as the Normalised Mean Face Coordinates (NMFC) image, and, along with the 2D gaze maps, we use it for conditioning the video renderer. When testing, we estimate the facial parameters of the source video, as done in training, while replacing the identity parameters of the source with the ones coming from the target video, and then render the result. 
\\
2- \textbf{Learning-based video rendering}. Given the conditional gaze and NMFC images, our  neural  network learns to translate its conditional input sequence to a highly realistic and temporally coherent output video, synthesising the target actor with exactly the same time-varying facial attributes in the source video. We adopt a GAN framework alongside an image discriminator and a multi-scale dynamics discriminator, which ensures that the generated video is realistic, temporally coherent and conveys the same dynamics of the target video. We further improve the visual quality of the mouth area, by designing a dedicated mouth discriminator. 

\section{Technical Setups and Results}
Our system has been tested on various machines. It only requires a commodity webcam for capturing the source actor interactively, and can generate synthetic videos in real time when run on a machine with a Graphics Processing Unit (GPU). Practically, the participant will set in front of our laptop and interactively drive the target person they choose to reenact before starting the demo. We provide some trained models on celebrities, politicians, athletes, etc. that the participant can drive with their faces. The only thing we require is a screen to connect our laptop to for aiding the visualisation. 




\bibliographystyle{ieeetr}
\bibliography{bibliography}

\end{document}